\documentclass[conference,a4paper]{IEEEtran}
\IEEEoverridecommandlockouts

\usepackage{cuted}  
\usepackage[hidelinks]{hyperref}
\usepackage[cmex10]{amsmath}
\usepackage{amssymb,amsfonts}
\usepackage{dblfloatfix}

\usepackage[ruled,vlined]{algorithm2e}
\usepackage{graphicx}
\graphicspath{{Figures/PDF/}{Figures/PNG/}}

\usepackage{booktabs}
\usepackage{siunitx}
\usepackage[numbers,compress]{natbib}
\usepackage{texnames}
\usepackage{bm,bbm}
\usepackage{orcidlink}
\usepackage{adjustbox}
\usepackage{longtable}
\usepackage{svg}
\usepackage{multirow}
\begin{document}

\title{\uppercase{Vision Transformer-Based Time-Series Image Reconstruction for Cloud-Filling Applications}
}

\author{	\IEEEauthorblockN{Lujun LI\orcidlink{0009-0007-8698-0685}}
	\IEEEauthorblockA{\textit{University of Luxembourg}\\
		1359 Kirchberg, Luxembourg\\
		lujun.li@uni.lu}
	\and
	\IEEEauthorblockN{Yiqun WANG\orcidlink{0000-0001-8459-8334}}
	\IEEEauthorblockA{\textit{University of Luxembourg}\\
		1359 Kirchberg, Luxembourg\\
		yiqun.wang@uni.lu}
	\and
	\IEEEauthorblockN{Radu STATE\orcidlink{0000-0002-4751-9577}}
	\IEEEauthorblockA{\textit{University of Luxembourg}\\
		1359 Kirchberg, Luxembourg\\
		radu.state@uni.lu}
}

\maketitle


\begin{abstract}

Cloud cover in multispectral imagery (MSI) poses significant challenges for early season crop mapping, as it leads to missing or corrupted spectral information. Synthetic aperture radar (SAR) data, which is not affected by cloud interference, offers a complementary solution, but lack sufficient spectral detail for precise crop mapping. To address this, we propose a novel framework, Time-series MSI Image Reconstruction using Vision Transformer (ViT), to reconstruct MSI data in cloud-covered regions by leveraging the temporal coherence of MSI and the complementary information from SAR from the attention mechanism. Comprehensive experiments, using rigorous reconstruction evaluation metrics, demonstrate that Time-series ViT framework significantly outperforms baselines that use non-time-series MSI and SAR or time-series MSI without SAR, effectively enhancing MSI image reconstruction in cloud-covered regions\footnote{Additional results are available at: \url{https://github.com/DobricLilujun/SMTS-VIT}. This paper has been accepted as a conference paper at the 2025 IEEE International Geoscience and Remote Sensing Symposium (IGARSS).
}.

\end{abstract}

\begin{IEEEkeywords}
Image Reconstruction, Vision Transformer, Time-series, MSI Images, SAR Images.
\end{IEEEkeywords}

\section{Introduction}
Accurate early crop mapping is critical for agricultural monitoring, food security assessment, and sustainable resource management. Multispectral imagery (MSI), with its rich spectral information, has been widely adopted for crop classification tasks \cite{wang2024cross,wang2024cross2}. However, one of the significant limitations of MSI is its susceptibility to cloud cover, which frequently obscures critical observations during the growing season. This challenge is particularly problematic for early-season crop mapping and prediction, where consistent temporal information is crucial to capture crop growth dynamics and making informed decisions. Synthetic aperture radar (SAR) data \cite{chan2008introduction}, on the other hand, is unaffected by cloud interference and provides reliable all-weather, day-and-night observations. However, SAR lacks the spectral details of MSI, which are essential for distinguishing between crop types. Current methods often combine SAR and MSI data for crop mapping, and cloud-obscured MSI regions are typically addressed with simple techniques like linear interpolation. However, these methods fail to capture the complex temporal and spectral patterns needed for accurate reconstruction under heavy cloud cover, limiting crop classification effectiveness in such scenarios.

Building upon recent developments, we propose a novel framework, Time-Series MSI Image Reconstruction using Vision Transformer (ViT) \cite{DBLP:journals/corr/abs-2010-11929}, to tackle the issue of cloud-induced data gaps in MSI. Our method leverages the temporal coherence of MSI and the complementary structural information from SAR to reconstruct cloud-obscured MSI regions. The contributions of this work are as follows.


\begin{itemize}

\item {Cloud-aware MSI reconstruction with MSI-SAR integration: A ViT model leverages temporal-spatial dependencies and integrates SAR data with MSI to reconstruct cloud-covered MSI, preserving spectral integrity and addressing individual limitations.}

\item {Comprehensive evaluation: Validated across diverse years, demonstrating that the Time-series ViT framework significantly outperforms methods using non-time-series MSI and SAR or time-series MSI without SAR, especially under severe cloud cover.}

\end{itemize}

The results of this study highlight the potential of integrating MSI and SAR data through ViT techniques for robust and accurate MSI reconstruction, paving the way for more reliable agricultural monitoring solutions in cloud-prone areas.

\section{Related Works}

Early crop mapping is heavily dependent on remote sensing data, particularly multispectral imagery (MSI), for its rich spectral information and ability to capture vegetation characteristics. Vegetation indices (VIs) derived from MSI, such as the normalized difference vegetation index (NDVI) \cite{huang2021commentary}, are widely used to monitor crop growth and differentiate crop types. However, MSI data is vulnerable to cloud cover, causing data gaps during critical phenological stages. Traditional clustering techniques based on MSI-derived VIs, as shown in \cite{wang2023early}, and advanced methods utilizing spectral features \cite{wang2024cross,wang2024cross2} have underscored MSI's importance despite its limitations. Synthetic aperture radar (SAR), unaffected by cloud interference, offers consistent all-weather observations and has been widely used to complement MSI. SAR's temporal features enhance mapping performance \cite{gella2021mapping}, while convolutional neural networks applied to SAR time series data \cite{han2023spatio} address its lack of spectral richness essential for distinguishing crop types.

Integrating MSI and SAR data has emerged as a promising approach to overcome the limitations of single data sources. Studies \cite{forkuor2014integration,metrikaityte2022digital,mazza2023synergic} demonstrate improved classification through data fusion, with temporal integration enhancing performance in complex agricultural systems \cite{zhu2022identification}. Deep learning methods \cite{liu2024enhanced} refine feature extraction and fusion for robust classification models, while machine learning approaches \cite{tufail2022machine} leverage multi-source temporal dynamics for improved accuracy. Despite these advancements, existing reconstruction methods for cloud-affected MSI, including linear interpolation-based approaches \cite{wang2024cross,wang2024cross2} and single-time fusion of MSI and SAR data for NDVI reconstruction \cite{mao2022cloud,meraner2020cloud}, remain inadequate for capturing temporal and spectral complexities. Innovative methods that fully utilize SAR's potential to reconstruct MSI data and address cloud-induced gaps are critical for scalable and accurate early-season crop mapping in cloud-prone regions \cite{zhu2022identification,liu2024enhanced,tufail2022machine}.

Recent works on time-series image processing with ViTs primarily employ separate patch embeddings for spatial and temporal dimensions \cite{zhang2024multiscalerestorationmissingdata, tarasiou2023vitssitsvisiontransformers}, followed by isolated processing of these dimensions with the addition of corresponding positional encodings. While this approach enables the model to gradually learn spatiotemporal features, it significantly increases model complexity \cite{DBLP:journals/corr/abs-2102-05095} due to the expanded number of spatiotemporal patches, making training more challenging. In contrast, our time-series ViT directly concatenates multichannel data in temporal order and uses a feature extractor for projection into a latent space. This design efficiently captures temporal dependencies, as demonstrated by its performance in image reconstruction experiments, especially under heavy cloud coverage.

\section{Study Areas and Data Preparation}

\begin{figure}[hbt]
	\centering
	\includegraphics[width=0.73\linewidth]{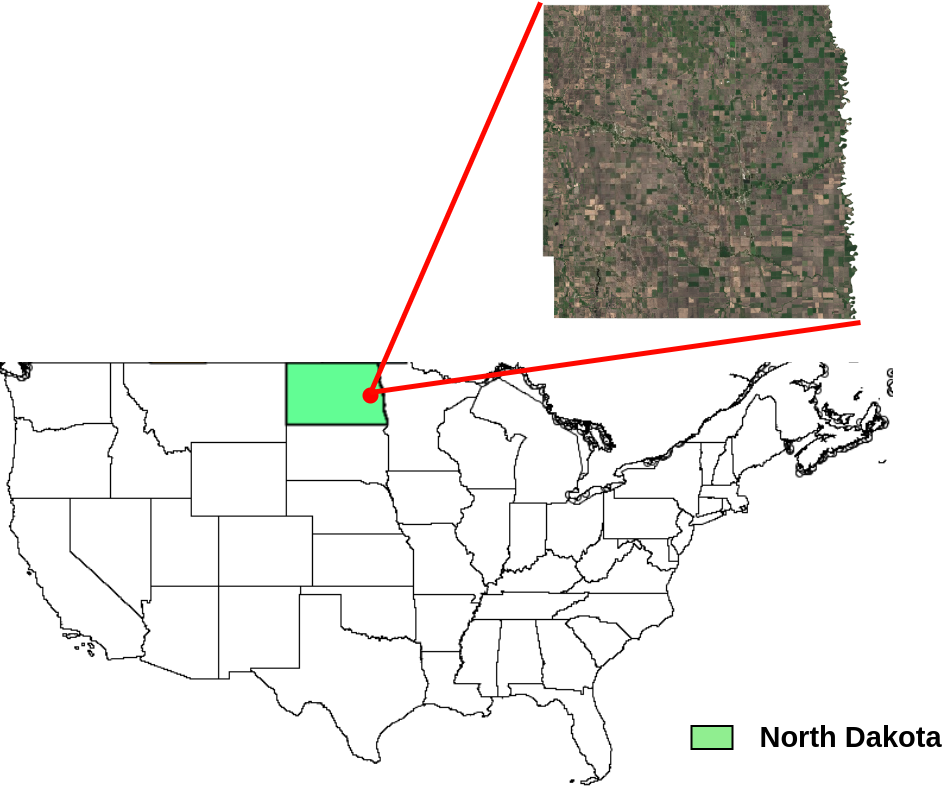}
	\caption{The Study Area Traill County located in North Dakota, the USA.}\label{fig:study_area}
\end{figure}

The study area selected for this research is Traill County, located in North Dakota, USA, as shown in Figure \ref{fig:study_area} which is an agriculturally significant region, making it an ideal testbed for image reconstruction and crop mapping. For remote sensing data, we used Sentinel-1 (SAR) \cite{torres2012gmes} and Sentinel-2 (MSI) \cite{main2017sen2cor} images. The Sentinel-1 data includes VV (vertical transmit, vertical receive) and VH (vertical transmit, horizontal receive) polarization bands. The Sentinel-2 data utilizes the following bands: B1 (Coastal Aerosol), B2 (Blue), B3 (Green), B4 (Red), B5 (Red Edge 1), B6 (Red Edge 2), B7 (Red Edge 3), B8 (Near Infrared, NIR), B8A (Red Edge 4), B11 (Shortwave Infrared 1, SWIR1), and B12 (Shortwave Infrared 2, SWIR2). The data collection period begins on May 1st and spans 60 days, covering critical stages of crop emergence and early growth. To ensure consistent spatial resolution across the data sets, all SAR and MSI images were reprojected onto a 30-meter grid, aligning with the resolution of the United States Department of Agriculture (USDA) Cropland Data Layer (CDL) \cite{boryan2011monitoring}.


To address data gaps in SAR and MSI acquisitions during their respective revisit periods, we implemented a data aggregation strategy. Although SAR and MSI nominally revisit every 6 and 5 days, respectively, practical acquisition limitations often resulted in missing data. To mitigate this, we combined data from two consecutive revisit periods, extending the SAR intervals to 12 days and the MSI intervals to 10 days. This approach yielded five SAR and six MSI images during the study period of 60 days, ensuring a consistent dataset while maintaining temporal resolution for analysis.

\section{Methodology}

\subsection{Problem Definition and Framework}
\subsubsection{Problem Definition}

The original remote sensing time series with cloud occlusion can be represented as \( X \in \mathbb{R}^{T \times C \times H \times W} \), where \( T \) denotes the length of the images in the time series, \( C \) is the number of channels at each time index and \( H \) and \( W \) represent the spatial dimensions of the region (height  and width). \( X(t, c, x, y) \) denotes the value of the pixel at position \( (x, y) \) in channel \( c \) at time \( t \). A cloud mask tensor \( M_{\text{cloud}} \in \mathbb{R}^{T \times C \times H \times W} \), as follows:

\begin{equation}
M_{\text{cloud}}(t, c, x, y) =
\begin{cases}
0 & \text{if } X(t, c, x, y) \text{ is non-cloud}, \\
1 & \text{if } X(t, c, x, y) \text{ is cloud}.
\end{cases}
\end{equation}


Assuming all channels have identical coverage at a given location and time \( t \), the missing value indicator tensor \( M_{\text{cloud}} \) is obtained from a known cloud detection algorithm Sentinel Hub's tool \cite{skakun2022cloud}. The target sequence \( Y \in \mathbb{R}^{T \times C \times H \times W} \) represents the actual data, while the prediction \( \hat{Y}=\text{Model}(X_i, M_{\text{cloud}}) \) is the reconstruction output of the ViT model. The model aims to minimize the loss function, ensuring that \( \hat{Y} \) closely matches the distribution of \( Y \) before cloud occlusion, reducing information loss. The loss function \( \mathcal{L}(\hat{Y}, Y) \) is detailed in Equation \ref{equa: loss}.




\subsubsection{General Framework}

\begin{figure}[ht]
    \centering
    \includegraphics[width=0.46\textwidth]{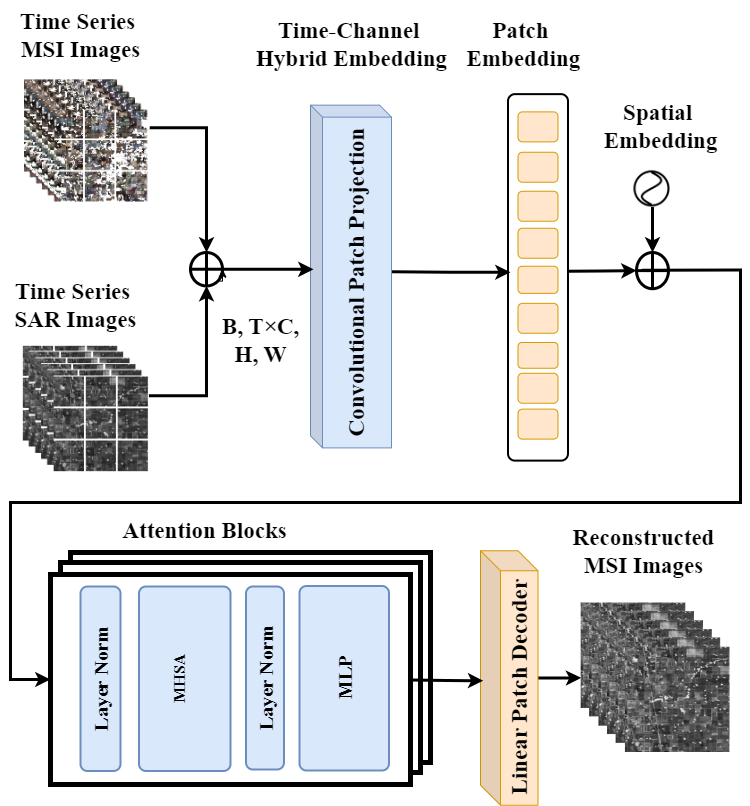} 
    \caption{The proposed Time-Series ViT reconstruction Structure}
    \label{fig:general strucutre}
\end{figure}

The framework proposed in this paper consists of Convolutional Patch Projection (CPP), a Multi-Head Self-Attention (MHSA) Encoder, and a Patch Decoder as shown in Fig. \ref{fig:general strucutre}. By freezing its pre-trained parameters, this framework can also be extended to early crop mapping in remote sensing images using transfer learning. The details will be discussed in the following sections. 

\subsection{Convolutional Patch Projection (CPP)}

Convolutional Neural Networks (CNNs) have been extensively utilized for feature map extraction. Building upon the foundational work of ViT \cite{DBLP:journals/corr/abs-2010-11929}, rather than employing separate temporal and spatial patch embeddings, we propose an integrated approach. Specifically, we merge the temporal channels of the input at the pixel level with MSI and SAR channels. This transforms the input tensor \( X \) into \( X' \), where the channel dimension \( C' \) is redefined as \( T \times C \), while the spatial dimensions \( H \) and \( W \) remain unchanged. Subsequently, a 2D convolution operation is applied to encode the input into patches. This process produces an embedded token vector for each pixel-level patch, wherein each token not only captures spatial pixel information but also encapsulates rich and intricate temporal information derived from the transformation.

\subsection{MHSA}
\textbf{Self-attention} is commonly used, known as the mechanism of attention, for relationships within different token embeddings \cite{vaswani2023attentionneed}. The basic idea of self-attention is to compute the attention score between each position and others using the query, key, and value. Multi-head attention enhances the model’s fitting power by processing multiple attention heads in parallel, then combining them for the output. This output is projected back to the embedding space to generate the patch embedding token after MHSA computation.

\subsection{Linear Patch Decoder \& Loss Function}



In this paper, we primarily introduce a multi-scale loss function that combines the Mean Squared Error (MSE) \cite{scikit-learn} loss and the Spectral Angle Mapper (SAM) \cite{dennison2004comparison,baldridge2009aster} loss. The MSE loss focuses on the performance of the global reconstruction, ensuring overall clarity and structural accuracy. The SAM loss emphasizes high-resolution details, particularly the consistency of spectral information. The multi-scale loss function integrates both MSE and SAM losses across multiple image scales, where linear interpolation is employed to transform the image to different scales (e.g., scales = [1, 0.5, 0.25]) and specified weights are assigned for each loss component:

\begin{equation}
\label{equa: loss}
\mathcal{L}_{\text{Multi-Scale}} = 
\sum_{s=1}^{S}
w_{\text{MSE}} \cdot 
\mathcal{L}_{\text{MSE}}^{(s)} + 
w_{\text{SAM}} \cdot 
\mathcal{L}_{\text{SAM}}^{(s)}
\end{equation}

where \( S \) is the number of scales, and \( w_{\text{MSE}} \) and \( w_{\text{SAM}} \) are the weights for MSE and SAM losses, respectively. For performance evaluation, we report the average performance metrics, including MSE, SAM, and Structural Similarity Index Measure (SSIM) \cite{1284395, 4775883}, computed between the generated and reference images. Additionally, we provide the Peak Signal-to-Noise Ratio (PSNR) \cite{1284395, 4775883} as a complementary measure. To facilitate qualitative assessment, we also present visual examples.




\begin{table*}[htbp]
\centering
\caption{Evaluation Results: The table presents S\-ViT, MTS-ViT, and SMTS-ViT models, each trained three times and evaluated on data from 2020 and 2021. "Val 2020" and "Val 2021" refer to remote sensing data from these years. The "\# Cloud" indicates the number of additional clouds, each with random textures, boundaries, and positions.}
\label{tab:table results} 
\begin{adjustbox}{max width=1.0\textwidth}
\begin{tabular}{cccccccccc}
\toprule
\textbf{Model} &
  \textbf{\#  Cloud} &
  \multicolumn{2}{c}{\textbf{MSE ↓}} &
  \multicolumn{2}{c}{\textbf{SAM (AVG) ↓}} &
  \multicolumn{2}{c}{\textbf{PSNR (AVG) ↑}} &
  \multicolumn{2}{c}{\textbf{SSIM (AVG) ↑}} \\
  \cmidrule(lr){1-1} \cmidrule(lr){2-2} \cmidrule(lr){3-4} \cmidrule(lr){5-6} \cmidrule(lr){7-8} \cmidrule(lr){9-10}
  
\multicolumn{1}{l}{} &
  \multicolumn{1}{l}{} &
  Val 2020 &
  Val 2021 &
  Val 2020 &
  Val 2021 &
  Val 2020 &
  Val 2021 &
  Val 2020 &
  Val 2021 \\
  \midrule
\multirow{3}{*}{S-ViT}    & 20  & 0.010 & 0.016 & 0.211 & 0.248 & 21.249 & 19.286 & 0.562 & 0.529 \\
                          & 30  & 0.014 & 0.014 & 0.247 & 0.260 & 19.730 & 19.969 & 0.485 & 0.500 \\
                          & 40  & 0.018 & 0.018 & 0.276 & 0.293 & 18.626 & 18.851 & 0.433 & 0.452 \\
                          \midrule
\multirow{3}{*}{MTS-ViT}  & 20  & 0.011 & 0.014 & 0.227 & 0.266 & 20.562 & 19.869 & 0.476 & 0.470 \\
                          & 30  & 0.009 & 0.012 & 0.204 & 0.245 & 21.534 & 20.667 & 0.487 & 0.474 \\
                          & 40  & 0.012 & 0.015 & 0.223 & 0.265 & 20.093 & 19.467 & 0.424 & 0.411 \\
                          \midrule
\multirow{3}{*}{SMTS-ViT} & 20  & 0.008 & 0.012 & 0.201 & 0.243 & 21.918 & 20.536 & 0.524 & 0.481 \\
                          & 30  & 0.008 & 0.012 & 0.195 & 0.241 & 21.978 & 20.490 & 0.520 & 0.474 \\
                          & 40  & 0.010 & 0.014 & 0.213 & 0.261 & 20.955 & 19.708 & 0.469 & 0.432 \\
                          \midrule
S-ViT (AVG)                     & - & 0.014 & 0.016 & 0.245 & 0.267 & 19.868 & 19.369 & 0.493 & \textbf{0.493} \\
\midrule
MTS-ViT (AVG)                   & - & 0.011 & 0.014 & 0.218 & 0.259 & 20.730 & 20.001 & 0.462 & 0.452 \\
\midrule
SMTS-ViT (AVG)                  & - & \textbf{0.009} & \textbf{0.013} & \textbf{0.203} & \textbf{0.248} & \textbf{21.617} & \textbf{20.245} & \textbf{0.504} & 0.462 \\
\bottomrule
\end{tabular}

\end{adjustbox}
\end{table*}

\section{Experiments}

\subsection{Experiments Setup}

We perform an extensive evaluation of our model across a range of testing scenarios. Specifically, we train the baseline model on datasets that exclude temporal sequences, employing a single-frame Vision Transformer (Single-ViT, SViT) configuration. In this model, the ViT is trained on individual MSI image inputs, without any temporal sequence information. Subsequently, we primarily train our model on datasets composed solely of MSI temporal sequences (MSI-Time-Series-ViT, MTS-ViT) and datasets that integrate both MSI and SAR temporal sequences (SAR-MSI-Time-Series-ViT, SMTS-ViT). Furthermore, we evaluated the performance of the model in image reconstruction tasks, providing empirical evidence of its efficacy in the 2020 and 2021 datasets.


\subsection{Datasets \& Settings}

Geospatial data from the northern United States (2020) was used, including 11-dimensional MSI and 2-dimensional SAR data. The dataset was split into 80\% for training and 20\% for validation, and then divided into smaller 60x60 pixel images for further training. To assess generalization, MSI and SAR data from the same regions in 2021 served as an additional validation set. Artificial cloud coverage was introduced during training using Gaussian smooth random noise. The size of the cloud (default $cloud\_size = 0.3$, i.e., the clouds start at 30\% of the image size) and the number (default $\# \text{Clouds} = 10$) were adjustable. The cloud masks were applied via element-wise multiplication to simulate coverage. The original dataset had 4. 7\% cloud coverage, which increased to 32.2\% after artificial simulation.
Key model parameters such as depth, number of MHSA heads, and patch size, which influence both resolution and computational complexity, were selected based on prior work and constraints of GPU resources \cite{DBLP:journals/corr/abs-2010-11929}. The experimental setup is: patch size = 5, depth = 6, attention heads = 8, embedding dimension = 64, and 100 training epochs. The learning rate is set to \(1 \times 10^{-4}\), and random seeds are fixed at 42 for reproducibility. Loss weights are \( w_{\text{MSE}} = w_{\text{SAM}} = 0.5 \). Cloud counts of 20, 30, and 40 are evaluated.

\subsection{Results}

\begin{figure}[!htbp]
    \centering
    \includegraphics[width=0.45\textwidth]{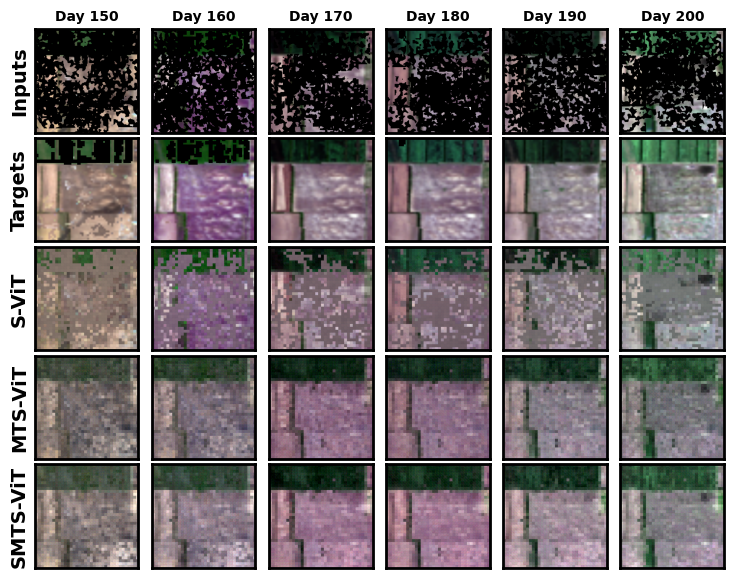} 
    \caption{The reconstructed images from the time-series input model are shown, with the x-axis representing the days of the year. "Inputs" refer to the cloud-augmented data, while "Targets" indicate the desired outputs (black pixels are real clouds). The remaining images display the results from the three models trained in this study, with a cloud count of 20 in this example.}
    \label{fig:png_example}
\end{figure}

Comparing the results of MTS-ViT and SMTS-ViT in 2020 with S-ViT (Table \ref{tab:table results}), both models show better performance on reconstruction tasks. SMTS-ViT achieves a 50\% improvement in overall MSE compared to S-ViT and a 20\% improvement over MTS-ViT. This highlights the effectiveness of our data augmentation and training framework, demonstrating the model's ability to efficiently infer missing image details in the time series. Combining with Figure \ref{fig:png_example}, we can observe that the reconstruction results of SMTS-ViT are visually more detailed and have a more consistent overall color tone. In contrast, the recovery with S-ViT exhibits numerous noise artifacts and abnormal color variations. A 4\% improvement for SMTS-ViT in PSNR and an 8\% improvement in SSIM comparing with MTS-ViT, showing that adding SAR data helps restore details and handle cloud occlusion. An increased number of clouds makes reconstruction harder for S-ViT, which struggles without prior and subsequent time series data, resulting in more guesswork and errors. Despite a slightly higher SSIM, S-ViT exhibited inferior visual quality and information recovery compared to MTS/SMTS-ViT, while overall performance in 2021 remained stable, demonstrating strong generalization.

\section{Conclusions and Future Works}


This paper presents SMTS-ViT, a ViT-based model for image reconstruction under cloud occlusion. SMTS-ViT combines temporal and MSI-SAR channels directly, using CPP patch embedding to encode spatial dimensions, removing the need for separating spatiotemporal embeddings. It captures both temporal and spatial relationships, excels at modeling temporal dependencies, and performs exceptionally well on the 2021 validation set for cloud-covered reconstruction tasks. Comparisons with S-ViT show its robustness, particularly under heavy cloud cover. Both metric and visual evaluations confirm SMTS-ViT’s superior performance in cloud reconstruction, significantly outperforming other models using MHSA. Future work will focus on validating the model for cross-domain crop prediction and exploring whether to retrain the ViT classifier or freeze its parameters while adding a classification layer to improve performance. Efforts will also be made to explore performance limits by following scaling laws with larger datasets and models. In addition, the impact of MSI-SAR channel combinations on crop prediction will be investigated.


\newpage

\small
\bibliographystyle{IEEEtranN}
\bibliography{references}

@article{mao2022cloud,
  title={Cloud removal based on SAR-optical remote sensing data fusion via a two-flow network},
  author={Mao, Ruihan and Li, Hua and Ren, Gaofeng and Yin, Zhangcai},
  journal={IEEE Journal of Selected Topics in Applied Earth Observations and Remote Sensing},
  volume={15},
  pages={7677--7686},
  year={2022},
  publisher={IEEE}
}

@inproceedings{mazza2023synergic,
  title={Synergic Use of SAR and Optical Data for Feature Extraction},
  author={Mazza, Antonio and Ciotola, Matteo and Poggi, Giovanni and Scarpa, Giuseppe},
  booktitle={IGARSS 2023-2023 IEEE International Geoscience and Remote Sensing Symposium},
  pages={2061--2064},
  year={2023},
  organization={IEEE}
}

@article{meraner2020cloud,
  title={Cloud removal in Sentinel-2 imagery using a deep residual neural network and SAR-optical data fusion},
  author={Meraner, Andrea and Ebel, Patrick and Zhu, Xiao Xiang and Schmitt, Michael},
  journal={ISPRS Journal of Photogrammetry and Remote Sensing},
  volume={166},
  pages={333--346},
  year={2020},
  publisher={Elsevier}
}

@article{skakun2022cloud,
  title={Cloud Mask Intercomparison eXercise (CMIX): An evaluation of cloud masking algorithms for Landsat 8 and Sentinel-2},
  author={Skakun, Sergii and Wevers, Jan and Brockmann, Carsten and Doxani, Georgia and Aleksandrov, Matej and Bati{\v{c}}, Matej and Frantz, David and Gascon, Ferran and G{\'o}mez-Chova, Luis and Hagolle, Olivier and others},
  journal={Remote Sensing of Environment},
  volume={274},
  pages={112990},
  year={2022},
  publisher={Elsevier}
}

@article{chan2008introduction,
  title={An introduction to synthetic aperture radar (SAR)},
  author={Chan, Yee Kit and Koo, Voon},
  journal={Progress In Electromagnetics Research B},
  volume={2},
  pages={27--60},
  year={2008},
  publisher={EMW Publishing}
}

@article{huang2021commentary,
  title={A commentary review on the use of normalized difference vegetation index (NDVI) in the era of popular remote sensing},
  author={Huang, Sha and Tang, Lina and Hupy, Joseph P and Wang, Yang and Shao, Guofan},
  journal={Journal of Forestry Research},
  volume={32},
  number={1},
  pages={1--6},
  year={2021},
  publisher={Springer}
}

@article{wang2023early,
  title={Early Crop Mapping Using Dynamic Ecoregion Clustering: A USA-Wide Study},
  author={Wang, Yiqun and Huang, Hui and State, Radu},
  journal={Remote Sensing},
  volume={15},
  number={20},
  pages={4962},
  year={2023},
  publisher={MDPI}
}

@article{wang2024cross,
  title={Cross Domain Early Crop Mapping using CropSTGAN},
  author={Wang, Yiqun and Huang, Hui and State, Radu},
  journal={IEEE Access},
  year={2024},
  publisher={IEEE}
}

@article{wang2024cross2,
  title={Cross Domain Early Crop Mapping with Label Spaces Discrepancies using MultiCropGAN},
  author={ Wang, Yiqun and Huang, Hui and State, Radu},
  journal={ISPRS Annals of the Photogrammetry, Remote Sensing and Spatial Information Sciences},
  volume={10},
  pages={241--248},
  year={2024},
  publisher={Copernicus Publications G{\"o}ttingen, Germany}
}

@article{gella2021mapping,
  title={Mapping crop types in complex farming areas using SAR imagery with dynamic time warping},
  author={Gella, Getachew Workineh and Bijker, Wietske and Belgiu, Mariana},
  journal={ISPRS journal of photogrammetry and remote sensing},
  volume={175},
  pages={171--183},
  year={2021},
  publisher={Elsevier}
}

@article{han2023spatio,
  title={Spatio-temporal multi-level attention crop mapping method using time-series SAR imagery},
  author={Han, Zhu and Zhang, Ce and Gao, Lianru and Zeng, Zhiqiang and Zhang, Bing and Atkinson, Peter M},
  journal={ISPRS Journal of Photogrammetry and Remote Sensing},
  volume={206},
  pages={293--310},
  year={2023},
  publisher={Elsevier}
}

@article{forkuor2014integration,
  title={Integration of optical and Synthetic Aperture Radar imagery for improving crop mapping in Northwestern Benin, West Africa},
  author={Forkuor, Gerald and Conrad, Christopher and Thiel, Michael and Ullmann, Tobias and Zoungrana, Evence},
  journal={Remote sensing},
  volume={6},
  number={7},
  pages={6472--6499},
  year={2014},
  publisher={MDPI}
}

@article{metrikaityte2022digital,
  title={Digital mapping of land cover changes using the fusion of SAR and MSI satellite data},
  author={Metrikaityte, Guste and Suziedelyte Visockiene, Jurate and Papsys, Kestutis},
  journal={Land},
  volume={11},
  number={7},
  pages={1023},
  year={2022},
  publisher={MDPI}
}

@article{zhu2022identification,
  title={Identification of soybean based on Sentinel-1/2 SAR and MSI imagery under a complex planting structure},
  author={Zhu, Mengqi and She, Bao and Huang, Linsheng and Zhang, Dongyan and Xu, Haifeng and Yang, Xiaoying},
  journal={Ecological Informatics},
  volume={72},
  pages={101825},
  year={2022},
  publisher={Elsevier}
}

@article{liu2024enhanced,
  title={Enhanced crop classification through integrated optical and SAR data: a deep learning approach for multi-source image fusion},
  author={Liu, Niantang and Zhao, Qunshan and Williams, Richard and Barrett, Brian},
  journal={International Journal of Remote Sensing},
  volume={45},
  number={19-20},
  pages={7605--7633},
  year={2024},
  publisher={Taylor \& Francis}
}

@article{tufail2022machine,
  title={A machine learning approach for accurate crop type mapping using combined SAR and optical time series data},
  author={Tufail, Rahat and Ahmad, Adeel and Javed, Muhammad Asif and Ahmad, Sajid Rashid},
  journal={Advances in Space Research},
  volume={69},
  number={1},
  pages={331--346},
  year={2022},
  publisher={Elsevier}
}

@article{torres2012gmes,
  title={GMES Sentinel-1 mission},
  author={Torres, Ramon and Snoeij, Paul and Geudtner, Dirk and Bibby, David and Davidson, Malcolm and Attema, Evert and Potin, Pierre and Rommen, Bj{\"O}rn and Floury, Nicolas and Brown, Mike and others},
  journal={Remote sensing of environment},
  volume={120},
  pages={9--24},
  year={2012},
  publisher={Elsevier}
}

@inproceedings{main2017sen2cor,
  title={Sen2Cor for sentinel-2},
  author={Main-Knorn, Magdalena and Pflug, Bringfried and Louis, Jerome and Debaecker, Vincent and M{\"u}ller-Wilm, Uwe and Gascon, Ferran},
  booktitle={Image and signal processing for remote sensing XXIII},
  volume={10427},
  pages={37--48},
  year={2017},
  organization={SPIE}
}

@article{boryan2011monitoring,
  title={Monitoring US agriculture: the US department of agriculture, national agricultural statistics service, cropland data layer program},
  author={Boryan, Claire and Yang, Zhengwei and Mueller, Rick and Craig, Mike},
  journal={Geocarto International},
  volume={26},
  number={5},
  pages={341--358},
  year={2011},
  publisher={Taylor \& Francis}
}

@misc{vaswani2023attentionneed,
      title={Attention Is All You Need}, 
      author={Ashish Vaswani and Noam Shazeer and Niki Parmar and Jakob Uszkoreit and Llion Jones and Aidan N. Gomez and Lukasz Kaiser and Illia Polosukhin},
      year={2023},
      eprint={1706.03762},
      archivePrefix={arXiv},
      primaryClass={cs.CL},
      url={https://arxiv.org/abs/1706.03762}, 
}

@misc{zhang2024multiscalerestorationmissingdata,
      title={Multi-scale Restoration of Missing Data in Optical Time-series Images with Masked Spatial-Temporal Attention Network}, 
      author={Zaiyan Zhang and Jining Yan and Yuanqi Liang and Jiaxin Feng and Haixu He and Li Cao},
      year={2024},
      eprint={2406.13358},
      archivePrefix={arXiv},
      primaryClass={cs.CV},
      url={https://arxiv.org/abs/2406.13358}, 
}

@misc{tarasiou2023vitssitsvisiontransformers,
      title={ViTs for SITS: Vision Transformers for Satellite Image Time Series}, 
      author={Michail Tarasiou and Erik Chavez and Stefanos Zafeiriou},
      year={2023},
      eprint={2301.04944},
      archivePrefix={arXiv},
      primaryClass={cs.CV},
      url={https://arxiv.org/abs/2301.04944}, 
}

@article{DBLP:journals/corr/abs-2010-11929,
  author       = {Alexey Dosovitskiy and
                  Lucas Beyer and
                  Alexander Kolesnikov and
                  Dirk Weissenborn and
                  Xiaohua Zhai and
                  Thomas Unterthiner and
                  Mostafa Dehghani and
                  Matthias Minderer and
                  Georg Heigold and
                  Sylvain Gelly and
                  Jakob Uszkoreit and
                  Neil Houlsby},
  title        = {An Image is Worth 16x16 Words: Transformers for Image Recognition
                  at Scale},
  journal      = {CoRR},
  volume       = {abs/2010.11929},
  year         = {2020},
  url          = {https://arxiv.org/abs/2010.11929},
  eprinttype    = {arXiv},
  eprint       = {2010.11929},
  bibsource    = {dblp computer science bibliography, https://dblp.org}
}

@article{DBLP:journals/corr/abs-2102-05095,
  author       = {Gedas Bertasius and
                  Heng Wang and
                  Lorenzo Torresani},
  title        = {Is Space-Time Attention All You Need for Video Understanding?},
  journal      = {CoRR},
  volume       = {abs/2102.05095},
  year         = {2021},
  url          = {https://arxiv.org/abs/2102.05095},
  eprinttype    = {arXiv},
  eprint       = {2102.05095},
  bibsource    = {dblp computer science bibliography, https://dblp.org}
}

@article{dennison2004comparison,
  title={A comparison of error metrics and constraints for multiple endmember spectral mixture analysis and spectral angle mapper},
  author={Dennison, Philip E and Halligan, Kerry Q and Roberts, Dar A},
  journal={Remote Sensing of Environment},
  volume={93},
  number={3},
  pages={359--367},
  year={2004},
  publisher={Elsevier}
}

@article{baldridge2009aster,
  title={The ASTER spectral library version 2.0},
  author={Baldridge, Alice M and Hook, Simon J and Grove, CI and Rivera, GJRSoE},
  journal={Remote sensing of environment},
  volume={113},
  number={4},
  pages={711--715},
  year={2009},
  publisher={Elsevier}
}

@ARTICLE{4775883,
  author={Wang, Zhou and Bovik, Alan C.},
  journal={IEEE Signal Processing Magazine}, 
  title={Mean squared error: Love it or leave it? A new look at Signal Fidelity Measures}, 
  year={2009},
  volume={26},
  number={1},
  pages={98-117},
  doi={10.1109/MSP.2008.930649}}

@ARTICLE{1284395,
  author={Zhou Wang and Bovik, A.C. and Sheikh, H.R. and Simoncelli, E.P.},
  journal={IEEE Transactions on Image Processing}, 
  title={Image quality assessment: from error visibility to structural similarity}, 
  year={2004},
  volume={13},
  number={4},
  pages={600-612},
  doi={10.1109/TIP.2003.819861}}

@article{scikit-learn,
  title={Scikit-learn: Machine Learning in {P}ython},
  author={Pedregosa, F. and Varoquaux, G. and Gramfort, A. and Michel, V.
          and Thirion, B. and Grisel, O. and Blondel, M. and Prettenhofer, P.
          and Weiss, R. and Dubourg, V. and Vanderplas, J. and Passos, A. and
          Cournapeau, D. and Brucher, M. and Perrot, M. and Duchesnay, E.},
  journal={Journal of Machine Learning Research},
  volume={12},
  pages={2825--2830},
  year={2011}
}

\end{document}